%% file: learnable_mpc.tex
\documentclass{article}

\usepackage[final]{corl_2022} 
\usepackage{graphicx}
\usepackage{amsmath}
\usepackage{amssymb}
\usepackage{amsthm}
\usepackage{enumitem}
\usepackage{bm}
\usepackage{subfigure}
\usepackage{paralist}
\usepackage{booktabs}
\usepackage{wrapfig}
\usepackage{sidecap}
\sidecaptionvpos{figure}{c}

\DeclareMathOperator*{\argmin}{arg\,min}
\usepackage{comment}
\usepackage{hyperref}

\newtheorem{theorem}{Theorem}[section]

\definecolor{green}{RGB}{11,155,13}

\title{Learning Model Predictive Controllers with Real-Time Attention for Real-World Navigation}

\author{
  Xuesu Xiao$^{*1,2}$, Tingnan Zhang$^{*3}$, Krzysztof Choromanski$^{*3}$, Edward Lee$^3$, \\ \textbf{Anthony Francis$^3$, Jake Varley$^3$, Stephen Tu$^3$, Sumeet Singh$^3$, Peng Xu$^3$, }\\\textbf{Fei Xia$^3$, Sven Mikael Persson$^2$, Dmitry Kalashnikov$^3$, Leila Takayama$^4$, }\\\textbf{Roy Frostig$^3$, Jie Tan$^3$, Carolina Parada$^3$, and Vikas Sindhwani$^3$}\\
  $^*$Equally Contributing Authors $^1$George Mason University \\$^2$Everyday Robots $^3$Robotics@Google $^4$Hoku Labs\\
}

\begin{document}
\maketitle
\vspace{-23pt}

\begin{abstract}
Despite decades of research, existing navigation systems still face real-world challenges when deployed in the wild, e.g., in cluttered home environments or in human-occupied public spaces. 
To address this, we present a new class of implicit control policies combining the benefits of imitation learning with the robust handling of system constraints from Model Predictive Control (MPC). 
Our approach, called Performer-MPC,\footnote{Project page: \url{https://performermpc.github.io}} uses a learned cost function parameterized by vision context embeddings provided by Performers---a low-rank implicit-attention Transformer. We jointly train the cost function and construct the controller relying on it, effectively solving end-to-end the corresponding bi-level optimization problem. 
We show that the resulting policy improves standard MPC performance by leveraging a few expert demonstrations of the desired navigation behavior in different challenging real-world scenarios. 
Compared with a standard MPC policy, Performer-MPC achieves \textbf{$>$40\%} better goal reached in cluttered environments and \textbf{$>$65\%} better on social metrics when navigating around humans. 
\end{abstract} 

\keywords{Model Predictive Control, Transformers, Performers, Highly-Constrained Navigation, Social Navigation, Learning-based Control} 

\begin{figure}[h]
\centering
\includegraphics[width=0.49\textwidth]{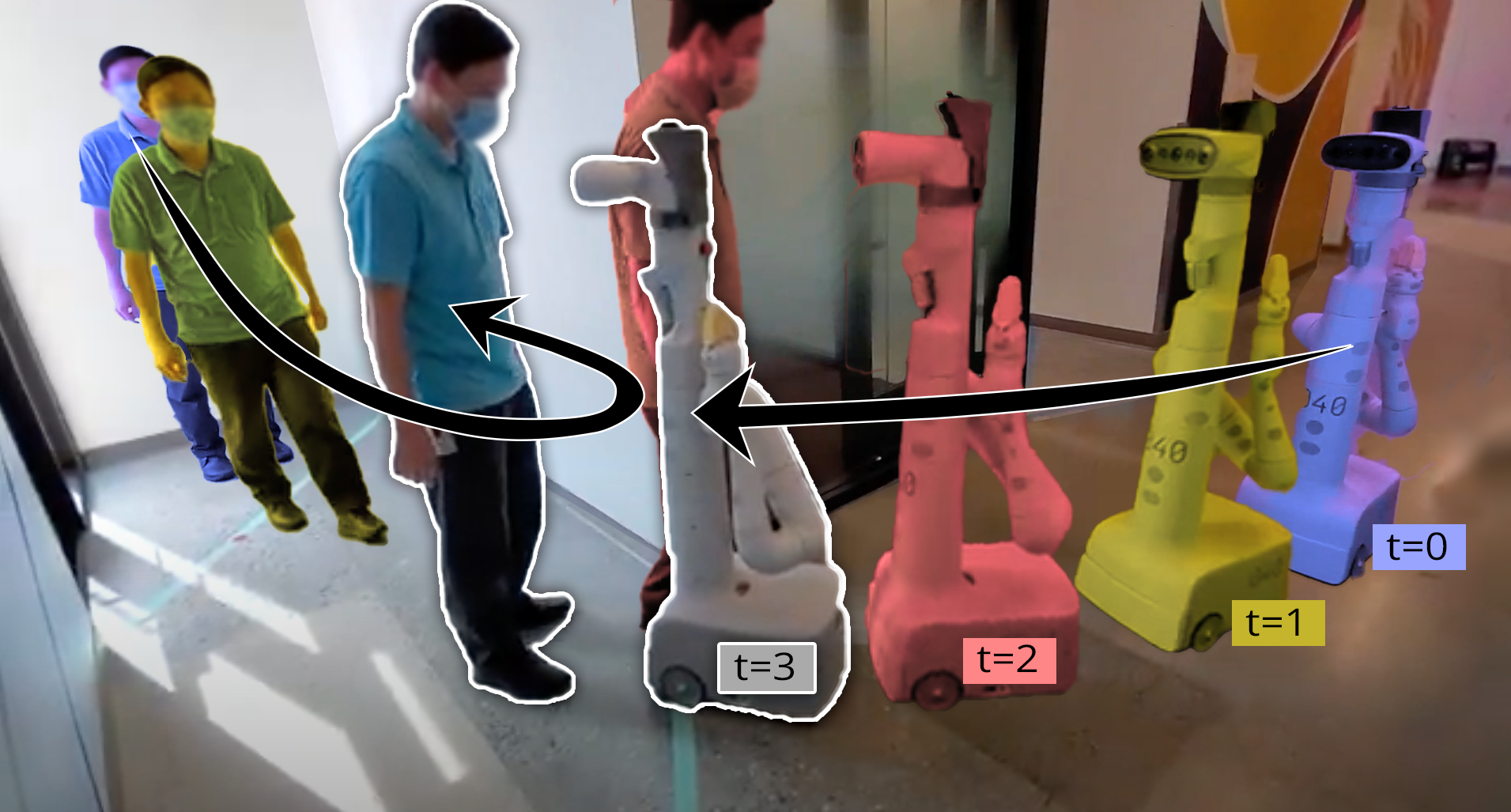}
\includegraphics[width=0.49\textwidth]{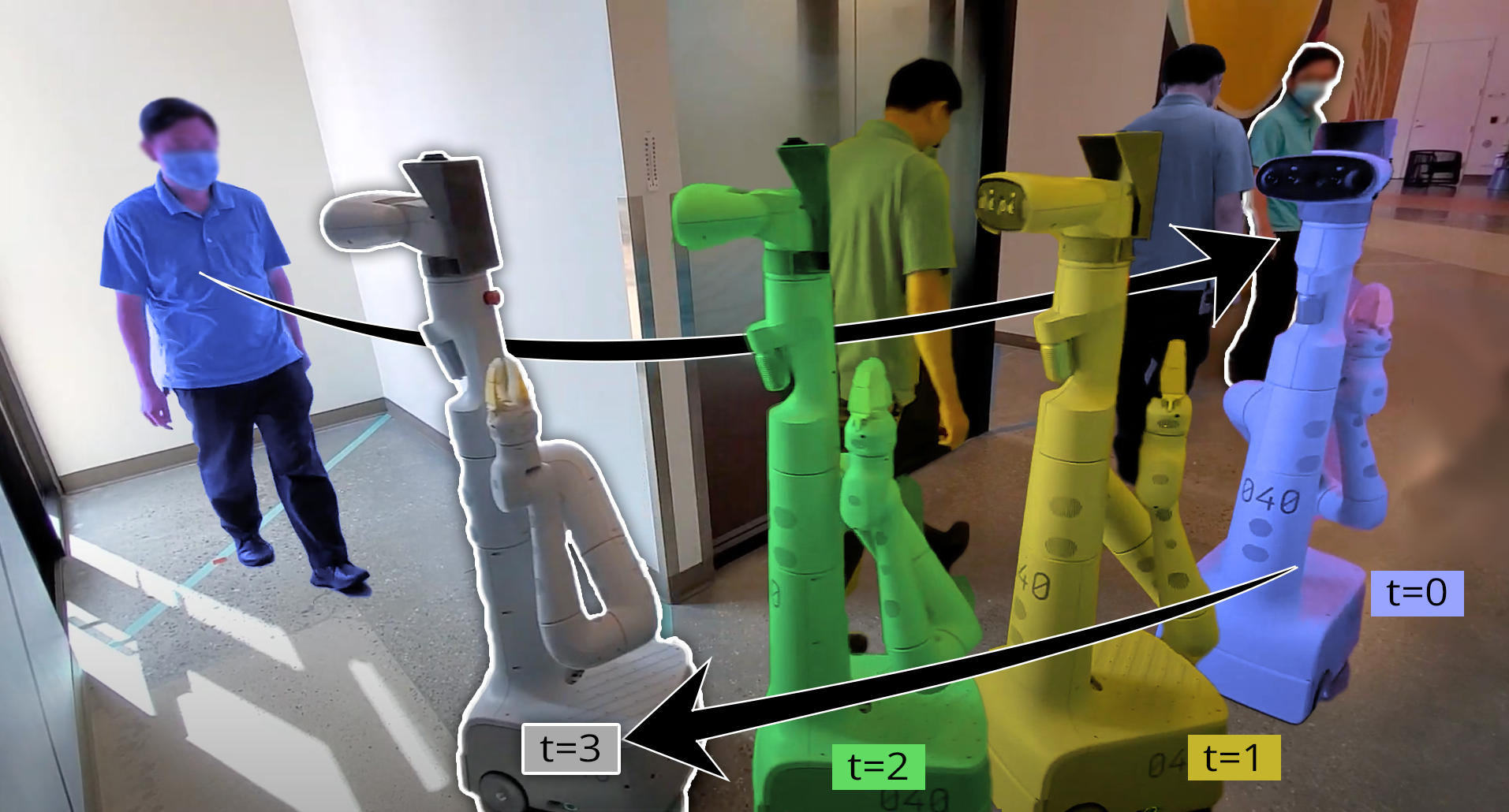}
\caption{Left: Standard  MPC efficiently cuts blind corners, forcing the human to back up; Right: Social Performer-MPC avoids cutting blind corners, enabling safe navigation around humans. }
\label{fig:blind_corner}
\end{figure}

\input{introduction}
\input{formulation}

\input{experiments}
\input{conclusion}


\clearpage


\bibliography{learnable_mpc}  

\input{supplementary}

\end{document}

%% file: introduction.tex
\vspace{-4mm}
\section{Introduction and Related Work}
\label{sec:intro}

Real-world robot deployment in human-centric environments, such as cluttered homes or crowded offices, remains an unsolved problem \cite{mavrogiannis21-social-survey, xiao2022autonomous}. These challenging situations require safe and efficient navigation through tight spaces, such as squeezing between coffee tables and couches, handling tight corners, doorways, untidy rooms, and so on. An equally critical requirement is to navigate in a manner that complies with unwritten social norms around humans.

Classical approaches using model-based control \cite{garcia1989model,mpc-1, mpc-nav,mpc-nonplanar} can already move robots from one point to another safely and reliably. However, when deploying these systems in the complex real world \cite{edr, scout, starship}, extensive engineering effort is required to construct world representations \cite{elfes1989tesselated, elfes1989using, dellaert2017factor}, model vehicle kinodynamics \cite{rabiee2019friction, tian2014control}, hand-craft cost functions \cite{lu2014layered, jaillet2010sampling}, fine-tune system parameters \cite{zheng2021ros}, and design backup planners to recover from stuck scenarios \cite{ros_move_base, mcnaughton2011motion, fox1997dynamic, quinlan1993elastic}. While providing verifiable guarantees, these cascaded components need to be hand-engineered before deployment based on the roboticist's best expectations of what would be encountered in the real world, and become cumbersome when in-situ modifications are necessary to enable adaptive behaviors \cite{zheng2021ros}. 

In contrast to these classical methods, machine learning enables robots to \emph{learn} these behaviors directly from data \cite{xiao2022motion}. 
End-to-end learning \cite{bojarski2016end, pfeiffer2017perception} is an appealing paradigm to reduce the engineering effort and cascading errors caused by separate components, but it usually requires extensive real-world training data or simulation with inevitably simplified human and environment representations. Most importantly, it lacks safety, optimality, generalizability, interpretability, and explainability, which are crucial for real robots moving around humans \cite{xiao2022motion, xu2021machine, xu2022benchmarking}. 
Therefore, researchers have looked at individually learning global planners \cite{yao2019following}, local planners \cite{faust2018prm, chiang2019learning, xiao2021toward, xiao2021agile, wang2021agile, liu2021lifelong}, and other navigation components including cost representations \cite{sikand2021visual, drews2017aggressive, wigness2018robot, kim2016socially}, kinodynamic models \cite{karnan2022vi, xiao2021learning}, and planner parameters \cite{teso2019predictive, bhardwaj2020differentiable, xiao2022appl, xiao2020appld, wang2021appli, wang2021apple, xu2021applr} to enable both better navigation performance, and also off-road \cite{bajracharya2009autonomous, bagnell2010learning} and social navigation \cite{mavrogiannis21-social-survey, mirsky2021prevention,tolani2021visual}. 

Both classical and learning-based methods have their merits. Model Predictive Control (MPC) \cite{garcia1989model,mpc-1, mpc-nav,mpc-nonplanar} enables synthesis of real-time feedback controllers for robots operating in real-world environments that satisfy given safety constraints, optimality criteria, and kinodynamic models. 
To get the best of both worlds, we design a class of \emph{Learnable-MPC} policies enabling robots to learn navigation behaviors in real-world use cases by combining the flexibility of learning from demonstrations with the optimality properties (e.g., collision-free, shortest path) of MPC solutions. 
Our framework can also been seen as a class of Implicit Behavior Cloning policies \cite{florence2022implicit} that are aware of real-world robot-environment and robot-human interactions. 
Our contributions in this paper are three-fold:
\begin{compactitem}
\item We augment the cost function of MPC with learnable components parameterized by rich Transformer-based latent embeddings of real-world context.  Transformer architectures \cite{vaswani, bert, lin} have produced stunning advances in language modeling~\cite{palm, gpt3, glam, gopher, megatron, lambda}, image generation \cite{styleswin, esser, mchen, zhaozhang, vitgan, transgan}, and multi-modal reasoning \cite{flamingo, clip, git, hu, polyvit, zeng2022socratic}. This indisputable success comes at a computational price in proportion to the massive number of parameters learned (e.g., $175$ billion for GPT-3~\cite{gpt3}) as well as quadratic scaling in input sequence length of the core attention modules of these models. By generating context-dependent quadratic costs using Performers \cite{performers}---a low-rank linear-attention Transformer, we demonstrate how we can embed powerful {\it pixel-to-pixel} attention mechanisms in MPC while crucially retaining real-time solutions on a CPU onboard a mobile robot.
\item Using distributed bilevel optimization with implicit differentiation mechanisms, we train navigation policies on expert demonstrations to handle difficult navigation scenarios, with data augmentation strategies to mitigate well-known distribution shift issues that frequently plague behavioral cloning and other imitation learning approaches. 
\item We demonstrate that our Performer-MPC outperforms its counterparts in real-world challenging navigation scenarios, including highly constrained and human-occupied environments. 
Performer-MPC learns to achieve \textbf{$>$40\%} better goal reached in highly constrained environments and \textbf{$>$65\%} better behavior as captured by social metrics defined in the Appendix when moving around humans in a social navigation pilot study.
\end{compactitem}

%% file: formulation.tex
\vspace{-3mm}
\section{Performer-MPC: Learnable MPC with Scalable Real-Time Attention}
\label{sec:lmpac}
\begin{figure}
\centering
\includegraphics[width=1\columnwidth]{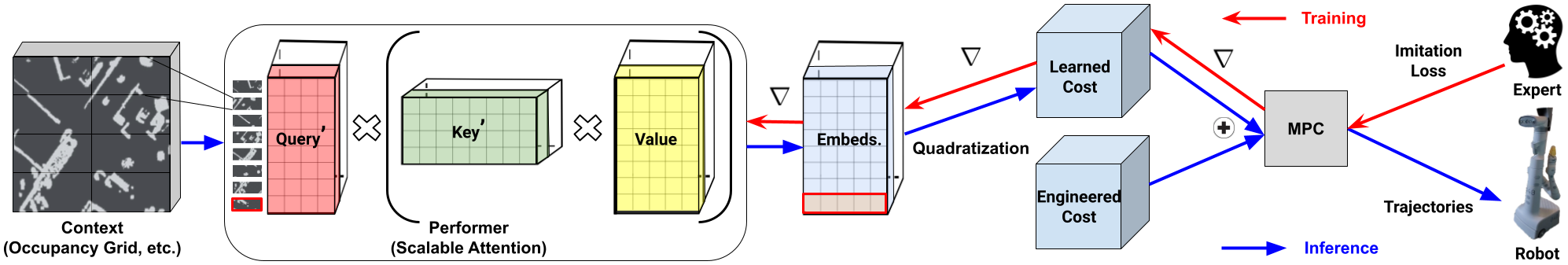}%
\caption{Overview of the Performer-MPC. The final latent embedding of the patch highlighted in red is used to construct context-dependent learnable cost. The backpropagation (red arrows) is through the parameters of the Transformer (see Sec. \ref{sec:attentive} for more details regarding Performers).}
\label{fig::overview}
\end{figure}

\vspace{-2mm}
In order to respond to dynamic uncertainty in the environment, the principal challenges of synthesis of model predictive controllers~\cite{borrelli2017predictive, rawlings2000tutorial,garcia1989model} are: (i) to construct cost functions that remain suitable across a wide variety of robot-environment situations, and (ii) to generate reliable solutions to the underlying trajectory optimization problems in real-time. This work focuses on the first challenge above, in particular, by eschewing the classical approach of hand-engineered cost functions, and adopting a learning-based inverse optimal control~\cite{levine2012continuous,mombaur2010human,boyd1994linear,palan2020fitting,amos2018differentiable,pauwels2014inverse} framework, where the sensory/visual context is used to induce {\it what MPC problem to solve in real-time} in order to generate actions. We first provide some details on training and inference of a learnable MPC framework, and then discuss the details of the learnable components; see Fig.~\ref{fig::overview} for an overview.

\subsection{Learnable Model Predictive Control}
\vspace{-2mm}
Let $C_0$ denote the current ``context", e.g., such as a list of tensors encoding for example RGB/D streams, force-torque readings, and proprioceptive states over a short time-window. Consider a {\it Learnable-MPC} feedback policy implicitly defined by solving the following \emph{parametric} optimal control problem at each time instant:
\vspace{-2mm}
\begin{subequations}
\begin{eqnarray}
&\argmin_{\mathbf{u}_0\ldots \mathbf{u}_{T-1}} &J_c(\mathbf{u}, \theta| C_0) =  \sum_{t=0}^{T-1} c(\mathbf{x}_t, \mathbf{u}_t, t; C_0, \theta) + c_T(\mathbf{x}_T; C_0, \theta), \\
&\mathrm{s.t.} &\mathbf{x}_{t+1} = f(\mathbf{x}_t, \mathbf{u}_t; \theta), \hspace{0.5cm} \mathbf{x}_0 = g(C_0, \theta) ~\textrm{given}. 
\end{eqnarray}
\label{eq:mpc}
\end{subequations}
Denote the optimizer as $\mathbf{u}^*(\mathbf{x}_0; C_0, \theta)$, and the corresponding optimal state sequence as $\mathbf{x}^*(\mathbf{x}_0; C_0, \theta)$. Here, $\theta$ are learnable parameters for stagewise and terminal cost neural networks $\{c, c_T\}$, $f$ the dynamics function, and $g$ current state estimator. While our framework generalizes to learning cost, dynamics, and state estimators simultaneously, in this paper we focus on the inverse optimal control setting: We study how the multi-layer self-attention cores of Transformers may be embedded in the \emph{cost networks} to handle sensor fusion {\it while retaining real-time speed expectations of MPC}. The dynamics function here corresponds to the differential drive dynamics of our robot.

While the MPC-structured policy can be trained using any flavor of Reinforcement Learning, real-world trial-and-error data is prohibitively expensive and a general reward function that captures all intricacies of different real-world scenarios is difficult to design. Therefore, we take an Imitation Learning approach where the robot has access to $N$ expert demonstrations. The MPC structure provides a form of a strong inductive bias for Imitation Learning, and can lead to improved data efficiency, robustness, and generalization. Denote  $\bar{\mathbf{u}}^i = (\bar{\mathbf{u}}^i_0,\ldots, \bar{\mathbf{u}}^i_{T_i-1})$ and $\bar{\mathbf{x}}^i = (\bar{\mathbf{x}}^i_0, \ldots, \bar{\mathbf{x}}^i_{T_i})$ as the control and state sequence for the $i^{th}$ demonstration snippet, with associated sensor context $C^i_0$, which can be extracted from offline planning or human teleoperation. We optimize $\theta$ as follows:
\vspace{-1mm}
\begin{eqnarray}
    \theta^\star &=& \arg\min_{\theta} \sum_{i=1}^N J_l(\mathbf{u}^{i*}(\theta) | \bar{\mathbf{u}}^i, \bar{\mathbf{x}}^i),
    \label{eq:theta_star}
    \end{eqnarray} where $J_l$ denotes total imitation loss that measures discrepancy between MPC-generated and expert state-control trajectories. We assume that $J_l$ also admits a stagewise and terminal decomposition using loss functions $l$ and $l_T$:
\vspace{-1mm}
\begin{subequations}
  \begin{eqnarray}   
    J_l(\mathbf{u}^{i*}(\theta) | \bar{\mathbf{u}}^i, \bar{\mathbf{x}}^i) &=& \sum_{t=0}^{T_i-1} l(\mathbf{x}^{i*}_t(\theta), \mathbf{u}^{i*}_t(\theta), t, \bar{\mathbf{x}}^i_t, \bar{\mathbf{u}}^i_t) + l_T(\mathbf{x}^{i*}_{T_i}(\theta), \bar{\mathbf{x}}^i_{T_i}), \\
    &&\mathbf{x}^{i*}_{t+1}(\theta) = f(\mathbf{x}^{i*}_t(\theta), \mathbf{u}^{i*}_t(\theta)), \hspace{0.5cm}\mathbf{x}^{i*}_0(\theta) = \bar{\mathbf{x}}^i_0.
\end{eqnarray}
\label{eq:sgd}
\end{subequations}
Above, $\mathbf{x}^{i*}, \mathbf{u}^{i*}$ is the MPC solution with cost parameters $\theta$, given context $C^i_0$ and initial state $\bar{\mathbf{x}}_0^i$.

\label{sec:bilevelopt}
\def\x{\mathbf{x}}
\def\u{\mathbf{u}}
\paragraph{Training via Bilevel Optimization: } The training optimization problem in~\eqref{eq:sgd} has embedded  the MPC optimization,~\eqref{eq:mpc}. Together, the two may be viewed as an instance of {\it bilevel optimization}, where the higher-level searches for the best  cost-network parameters $\theta$ via imitation loss minimization, while the lower-level  synthesizes the optimal predictive control sequences given fixed $\theta$. To use stochastic gradient descent for the higher-level problem, we need the gradient of $J_l$ with respect to $\theta$ evaluated at a control sequence $\mathbf{u}^{*}(\theta_k)$ where $\theta_k$ denotes the parameters during the current iterate $k$ during training. This quantity decomposes as a vector-Jacobian product (VJP),
\begin{equation}
 \nabla_{\theta} J_l(\mathbf{u}^{*}(\theta_k) | \bar{\mathbf{u}}^i, \bar{\mathbf{x}}^i) = \nabla_{\mathbf{u}} J_l(\mathbf{u}^{*}(\theta_k) | \bar{\mathbf{u}}^i, \bar{\mathbf{x}}^i)^T  \partial_\theta \mathbf{u}^*(\theta_k). \label{eq:vjp1}
\end{equation}

The first term in the product on the right hand side is the gradient of the total imitation loss, which can be efficiently computed using the Adjoint method in Optimal Control~\cite{bertsekas1997nonlinear} thanks to its stagewise structure. 
The second term is the sensitivity of the MPC solution with respect to parameters, which may be efficiently computed using the Implicit Function Theorem (IFT), as featured in several works exploring differentiable ``optimization layers"~\cite{lorraine2020optimizing,agrawal2019differentiable,agrawal2019learning,blondel2021efficient}, see Appendix for details.

\paragraph{MPC Solver: } We use a second order Gauss-Newton trajectory optimizer~\cite{dunn1989efficient,wright1990solution} called Iterative LQR (iLQR)~\cite{li2004iterative} with line searches inspired by Differential Dynamic Programming (DDP)~\cite{murray1984differential,dunn1989efficient}. At each iteration, iLQR quadratizes the cost and linearizes the dynamics to compute the search direction by solving a time-varying LQR (TVLQR)~\cite{bertsekas1997nonlinear} problem. Upon convergence, a single additional LQR solve suffices for computing $\partial_{\theta} \mathbf{u}^*(\theta)$ in~\eqref{eq:vjp1}.

\paragraph{Policy: } While one may use the first component of the optimal solution for problem~\eqref{eq:mpc}, i.e., $\u^*_0(\x_0; C_0, \theta)$, as the policy map, we noted better performance by leveraging a secondary (non-learnable) MPC problem similar to~\eqref{eq:mpc}, featuring a ``tracking" objective w.r.t. the solution of the learnable MPC problem. The details of this ``tracking MPC" problem are provided in the Appendix. 

\subsection{Attentive Cost Functions for Learnable MPC}
\label{sec:attentive}
We adopt an inverse optimal control framework for learnable MPC whereby only the cost function is learnable. We structure this cost as the sum of a user-engineered function and a context-dependent quadratic, parameterized by an embedding matrix $\mathbf{P}$ and vector $\mathbf{q}$ (described in more detail below):
\begin{equation}
    c(\mathbf{x}, \mathbf{u}, t; C_0, \theta) = \bar{c}(\mathbf{x}, \mathbf{u}, t) +  \left[\begin{array}{c}\mathbf{x}\\\mathbf{u}\end{array}\right]^T \mathbf{P}^T(C_0, \theta)\mathbf{P}(C_0, \theta) \left[\begin{array}{c}\mathbf{x}\\\mathbf{u}\end{array}\right] + \mathbf{q}(C_0, \theta)^T  \left[\begin{array}{c}\mathbf{x}\\\mathbf{u}\end{array}\right]. \label{eq:quadratic_cost}
\end{equation}

Here, $\bar{c}$ refers to the hand-designed cost function (see Appendix), appended to a  Transformer-backed cost model that attends to the current context $C_0$ to generate residual quadratic cost terms
for MPC to optimize. This structure removes the computational cost of repeated quadratization of a large network in the iLQR solver. Furthermore, since the residual cost is convex and well-conditioned, crude MPC solutions can generate reliable descent directions for the higher-level optimizer, even though applying IFT in gradient computation assumes the MPC solution is precisely a local minimum. We next describe the details of the embeddings $\mathbf{P}$ and $\mathbf{q}$.

\input{performers-compressed}

%% file: performers-compressed.tex
\paragraph{Generating Context-Dependent Transformer Embeddings:}
We outline a general Transformer-based backend for learnable-MPCs which leads to Performer-MPCs. 
The backend maps the current contexts $C_0$ into a latent embedding which can be reshaped into the matrices $\mathbf{P}$ and $\mathbf{q}$ to support the quadratic parameterization of Eqn.~\ref{eq:quadratic_cost}. For concreteness, let $C_0$ be an image frame, i.e., the occupancy grid in the robot frame. 
As in Vision Transformer architectures \cite{vit-1, vit-2, vit-3}, each frame is first independently pre-processed by a convolution layer, and then flattened to a sequence. Each  element (token) of the sequence corresponds to a different patch of the original frame which is then enriched with positional encodings. The length $L$ of this sequence is a patch-size hyper-parameter. The preprocessed input is then fed to regular attention and MLP layers. 
The final embedding of one of the tokens is chosen as a latent representation of the entire context $C_{0}$ to parameterize the learnable cost (e.g., via de-vectorization to $\mathbf{P}$ and $\mathbf{q}$ as in Eqn. \ref{eq:quadratic_cost}). Even though we take the final embedding of a single token, it contains signal from all the tokens since attention mixes information across tokens.

The attention used in regular Transformer architectures \citep{vaswani} linearly projects tokens' embeddings (via trainable transformations) into three matrices, $\mathbf{Q}, \mathbf{K}, \mathbf{V} \in \mathbb{R}^{L \times d}$, called \textit{queries}, \textit{keys} and \textit{values} respectively. The output of the attention is then defined as:
\begin{equation}
\label{eq:attnorm}
    \mathrm{Att}(\mathbf{Q}, \mathbf{K}, \mathbf{V}) = \mathbf{D}^{-1} \mathbf{A} \mathbf{V}, \quad 
    \mathbf{A} = \exp ( \mathbf{Q} \mathbf{K}^\top / \sqrt{d}), \quad \mathbf{D} = \mathrm{diag} ( \mathbf{A} \mathbf{1}_L ), 
\end{equation}    
where $\mathbf{A} \in \mathbb{R}^{L \times L}$ is called the \textit{attention matrix}. In the above formula, $\exp (\cdot)$ is applied elementwise, $\mathbf{1}_L$ is the all-ones vector of length $L$, and $\mathrm{diag} (\cdot)$ is a diagonal matrix with the input vector as the diagonal. Space and time complexity of computing Eqn.~\ref{eq:attnorm} are:  $O(L^{2}+Ld)$ and $O(L^2 d)$ respectively, because $\mathbf{A}$ has to be explicitly stored.  Quadratic time and space complexity in the number of patches makes this approach prohibitive for real-time robotic navigation. To address this, we apply a class of low-rank implicit linear-attention Transformer architectures, called \textit{Performers} \cite{performers}.

Performers interpret attention matrices $\mathbf{A} \in \mathbb{R}^{L \times L}$
as kernel matrices, i.e., $\mathbf{A}(i,j)$ is defined as: $\mathbf{A}(i,j) = \mathrm{K}(\mathbf{q}_{i},\mathbf{k}_{j})$, with $\mathbf{q}_{i}/\mathbf{k}_{j}$ standing for the $i^{th}/j^{th}$ query/key row-vector in $\mathbf{Q}/\mathbf{K}$ and kernel $\mathrm{K}:\mathbb{R}^{d} \times \mathbb{R}^{d} \rightarrow \mathbb{R}_{+}$ defined for the (randomized) mapping $\phi: \mathbb{R}^{d} \rightarrow \mathbb{R}^{m}$ (for $m >0$) as:
\begin{equation}
\label{kernel-def}
\mathrm{K}(\mathbf{x}, \mathbf{y}) = \mathbb{E}[\phi(\mathbf{x})^{\top}\phi(\mathbf{y})].
\end{equation}
We call $\phi(\mathbf{v})$ a \textit{(random) feature map} for $\mathbf{v} \in \mathbb{R}^{d}$. 
For $\mathbf{Q}^{\prime},\mathbf{K}^{\prime} \in \mathbb{R}^{L \times m}$ with rows given as $\phi(\mathbf{q}_{i})$ and $\phi(\mathbf{k}_{i})$ respectively,
Eqn.~\ref{kernel-def} leads directly to the efficient attention mechanism of the form:
\begin{equation}
    \widehat{\mathrm{Att}} (\mathbf{Q}, \mathbf{K}, \mathbf{V}) = \widehat{\mathbf{D}}^{-1} (\mathbf{Q}^{\prime}((\mathbf{K}^{\prime})^{\top} \mathbf{V})), \quad 
    \quad \widehat{\mathbf{D}} = \mathrm{diag} (\mathbf{Q}^{\prime}((\mathbf{K}^{\prime})^{\top} \mathbf{1}_L) ). \label{attention}
\end{equation} 
Here $\widehat{\mathrm{Att}}$ stands for the approximate attention and the parentheses indicate the order of computations. Such a mechanism is characterized by space complexity of $O(Lm + Ld + md)$ and time complexity of $O(Lmd)$ as opposed to $O(L^{2}+Ld)$ and $O(L^{2}d)$ of the regular attention mechanism.  Different mappings $\phi$ give rise to different Performer variants. Two most popular ones \cite{performers} are:
\begin{align}
\begin{split}
\phi_{\mathrm{exp}}(\mathbf{x}) &\overset{\mathrm{def}}{=}\frac{1}{\sqrt{m}}\mathrm{exp}(-\frac{\|\mathbf{x}\|_{2}^{2}}{2})\left(\exp(\omega_{1}^{\top}\mathbf{x}), \dots ,\exp(\omega_{m}^{\top}\mathbf{x}) \right) \textrm{ and, }\\
\phi_{\mathrm{relu}}(\mathbf{x}) &\overset{\mathrm{def}}{=} \frac{1}{\sqrt{m}}(\mathrm{ReLU}(x_{1}), \dots ,\mathrm{ReLU}(x_{d})) \textrm{ (with $m=d$)},
\end{split}
\end{align}
for $\omega_{1},\dots,\omega_{m} \sim \mathcal{N}(0,\mathbf{I}_{d})$ and
we will refer to the corresponding Performers as \textit{Performer-Exp} and \textit{Performer-ReLU} respectively. Mapping $\phi_{\mathrm{exp}}$ provides unbiased estimation of the softmax attention-kernel from Eqn. \ref{eq:quadratic_cost}, but requires random projections, whereas $\phi_{\mathrm{relu}}$ defines weaker attention-kernel, but does not use random projections and leads to the fastest Performer variant. In Section \ref{sec:perablations} and the Appendix, we provide a comprehensive speed benchmark for Performer-MPC variations, demonstrating Pixel-to-Pixel attention at real-time speeds. 

%% file: experiments.tex
\section{Experiments}

\label{sec:exps}

\definecolor{rmpc}{RGB}{255, 0, 0}
\definecolor{ep}{RGB}{30, 119, 180}
\definecolor{pmpc}{RGB}{36, 156, 36}
\definecolor{demo}{RGB}{102, 102, 102}
\definecolor{train}{RGB}{0, 121, 185}
\definecolor{test}{RGB}{255, 117, 0}
\definecolor{goal}{RGB}{0, 0, 255}

\begin{figure}
\centering
\includegraphics[width=1\columnwidth]{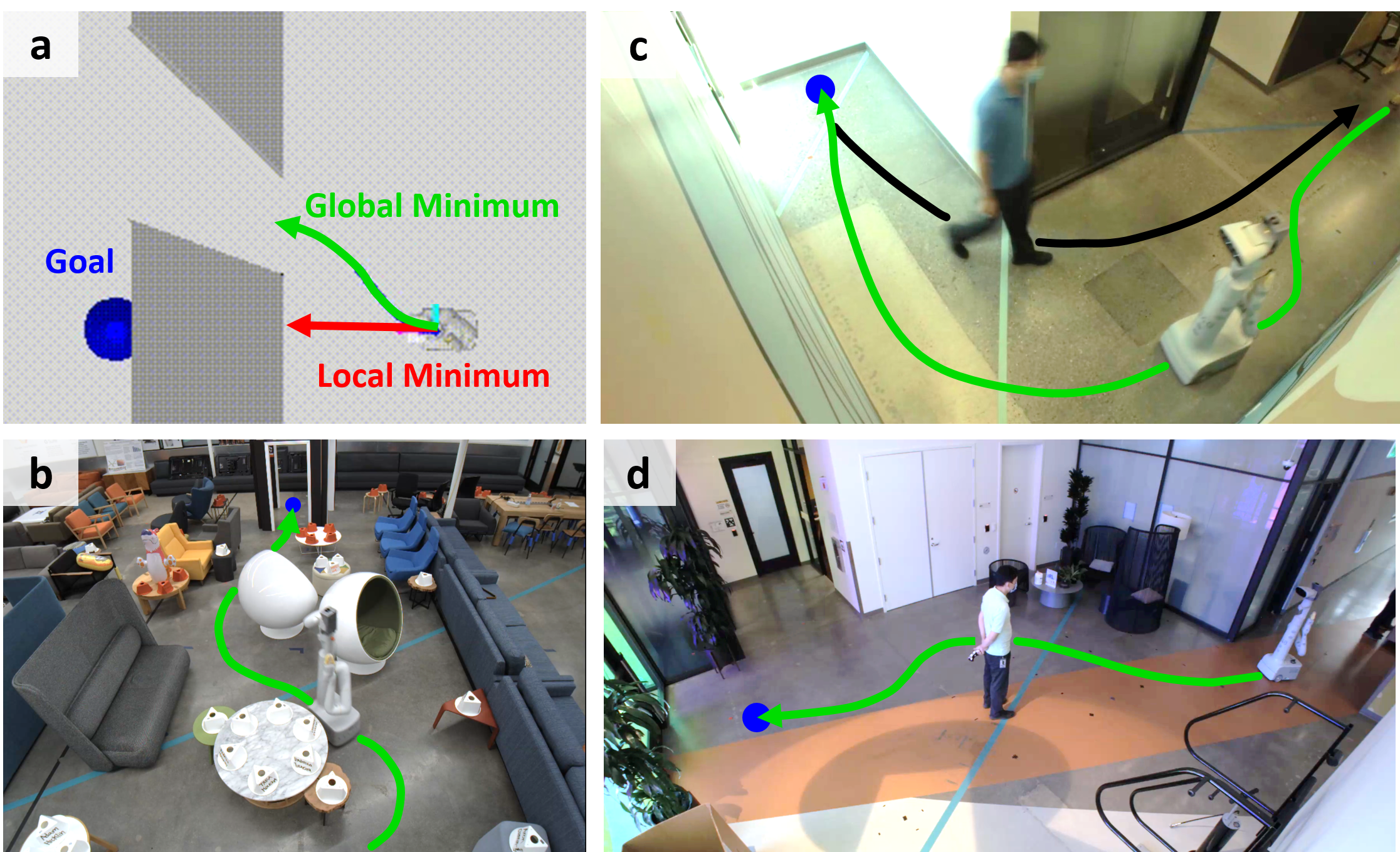}%
\label{fig::door_obstacle_social}
\caption{Experiment Scenarios: (a) Learning to avoid local minima during doorway traversal, (b) to maneuver through highly constrained spaces, and (c) to enable socially compliant behaviors for blind corner and (d) pedestrian obstruction interactions.}
\label{fig::experiment}
\end{figure}

Our \texttt{Performer-MPC} is tested on a differential-drive wheeled robot, which has a 3D LiDAR in the front, and depth sensors mounted on its head (see Fig.~\ref{fig:blind_corner}). 
It is a three-layer Performer-ReLU model with $\mathrm{mlpdim}=64$ and one head. 
Our policies are trained on four TPUs and then deployed on a CPU onboard the robot. We use the differentiable Iterative LQR implementation of \texttt{trajax} \cite{trajax2021github} for MPC training and inference. 
Please refer to the Appendix for more details. 

\texttt{Performer-MPC} is compared with two baselines, a regular MPC policy (\texttt{RMPC}) without the learned cost components, and an Explicit Policy (\texttt{EP}) that predicts a reference/goal state using the same Performer architecture, but without being coupled to the MPC structure. The control action (i.e., final policy output) is implicitly defined via the solution of the ``tracking MPC" problem (see previous section), where the reference trajectory is generated by one of \texttt{RMPC}, \texttt{EP}, or \texttt{Performer-MPC}.

\begin{figure}[t]
\centering
\includegraphics[width=1\textwidth]{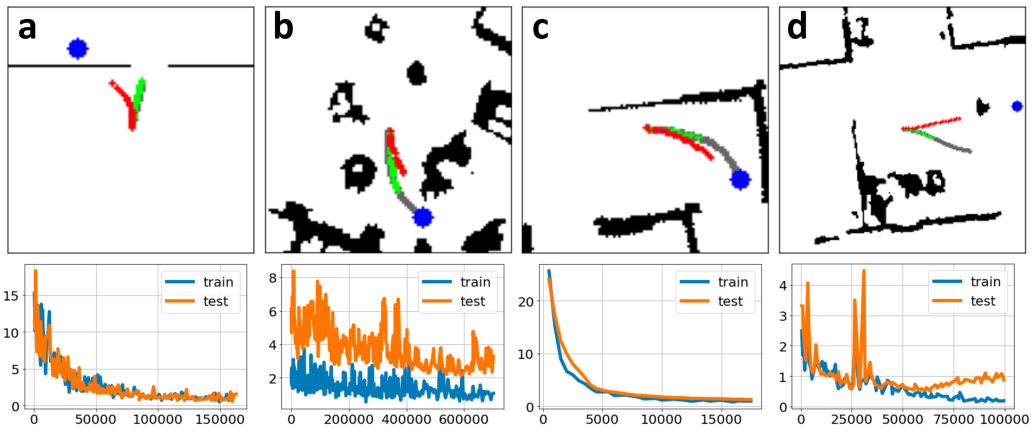}%
\caption{Top: Visualization of test data examples in the (a) doorway traversal, (b) highly constrained obstacle course, (c) blind corner, and (d) pedestrian obstruction scenarios. \textcolor{pmpc}{\texttt{Performer-MPC}} trajectories aiming at the \textcolor{goal}{goal} are always closer to the \textcolor{demo}{expert demonstrations} compared to the \textcolor{rmpc}{\texttt{RMPC}} trajectories. Bottom: \textcolor{train}{Train} and \textcolor{test}{test} curves, where the vertical axis represents loss values (Hausdorff distance to the expert trajectories) and horizontal axis represents training steps. }
\label{fig::test_data}
\end{figure}

We evaluate our method in four scenarios, one in simulation and three in the real world (Fig.~\ref{fig::experiment}). For each scenario, the learned policies (\texttt{EP} and \texttt{Performer-MPC} ) are trained with demonstrations specifically collected for that scenario. To address the distribution shift issue, we not only collect positive examples where the robot is driving smoothly with the intended behavior, but also start the robot in randomly selected ``disadvantage locations'' (e.g., near-collision situations), and steer the robot to recover from them. For more data collection details please refer to the Appendix. 
We visualize the planning results of \texttt{Performer-MPC} (\textcolor{pmpc}{green}) and \texttt{RMPC} (\textcolor{rmpc}{red}) along with expert demonstrations (\textcolor{demo}{grey}) in the top half of Fig. \ref{fig::test_data} and the train and test curves in the bottom half. 

\subsection{Experimental Results}

\paragraph{Learning to Avoid Local Minima: }
We first evaluate our method in a simulated doorway traversal scenario (Fig.~\ref{fig::experiment}a). 100 start and goal pairs are randomly sampled from opposing sides of the wall. A planner guided by a greedy cost function often leads the robot to a local minimum, i.e., getting stuck at the closest point to the goal on the other side of the wall. Although such a problem can be mitigated by using a global planner, we use this as a test case to showcase the learning results. We generate 2000 expert demonstrations using an off-line iLQR planner \cite{li2004iterative}, which iteratively solves for intermediate way points provided by a Dijkstra's global planner. Using these off-line demonstrations, \texttt{\texttt{Performer-MPC}} learns a cost landscape that steers the robot towards the doorway, even if it must veer away from the goal and travel further. \texttt{Performer-MPC}  passes the doorway in 86 out of 100 trials while \texttt{RMPC} only passes 24 out of 100. 

\vspace{-2mm}

\begin{figure}[tb]
\centering
\begin{minipage}{.65\textwidth}
\includegraphics[width=1\textwidth]{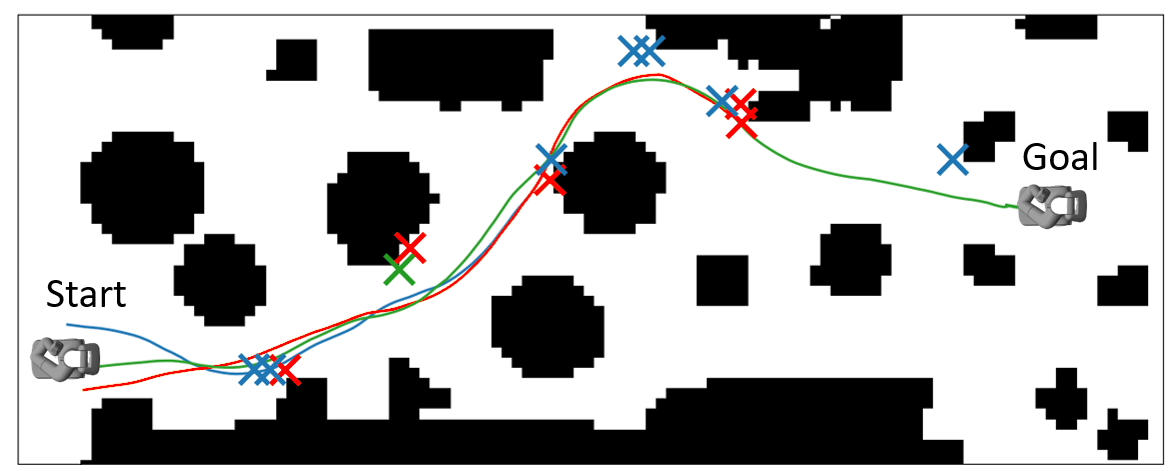}
\end{minipage}
\begin{minipage}{.34\textwidth}
\begin{tabular}{ccc}\\\toprule  
& \scriptsize SR & \scriptsize CP $\pm$ VAR\\ \midrule
\scriptsize \textcolor{rmpc}{\texttt{RMPC}} & \textcolor{rmpc}{\scriptsize 5/10} & \textcolor{rmpc}{\scriptsize $69\pm 13\%$} \\ 
\textcolor{ep}{\scriptsize \texttt{EP}} & \textcolor{ep}{\scriptsize 3/10} & \textcolor{ep}{\scriptsize $62\pm 15\%$} \\ 
\textcolor{pmpc}{\scriptsize \texttt{Performer-MPC}} & \textcolor{pmpc}{\scriptsize \bm{$9/10$}} & \textcolor{pmpc}{\scriptsize \bm{$92\pm 6\%$}}\\ \bottomrule
\end{tabular}
\end{minipage}
\caption{A $4.5\times 10$ $\textrm{m}^2$ obstacle course with policy trajectories and failure locations indicated by crosses for \textcolor{rmpc}{\texttt{RMPC}}, \textcolor{ep}{\texttt{EP}}, and \textcolor{pmpc}{\texttt{Performer-MPC}}.}
\label{fig:obstacle}
\end{figure}

\paragraph{Learning Highly-Constrained Maneuvers: }
We next test our method in a challenging real-world scenario---a cluttered home/office setting where the robot must perform sharp, near-collision maneuvers (Fig.~\ref{fig::experiment}b). A global planner provides coarse way points for the robot to follow. Each policy is run ten times and we report Success Rate (SR) and average Completion Percentage (CP) with variance (VAR) of the  obstacle course that the robot is able to traverse without collisions or getting stuck (Fig.~\ref{fig:obstacle}). \texttt{Performer-MPC} outperforms both \texttt{RMPC} and \texttt{EP} in SR and CP.

\begin{figure}[b]
\vspace{-5mm}
\centering
  \centering
  \includegraphics[width=0.49\linewidth]{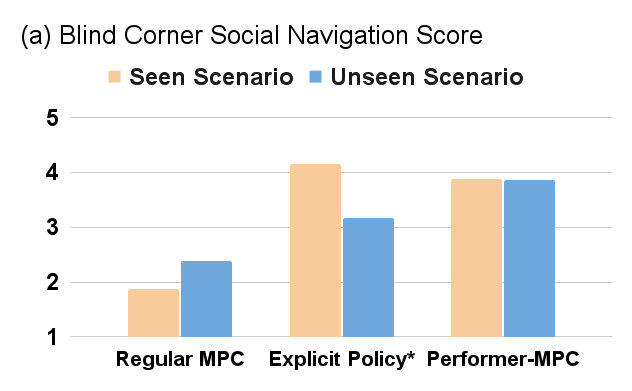}
  \centering
  \includegraphics[width=0.49\linewidth]{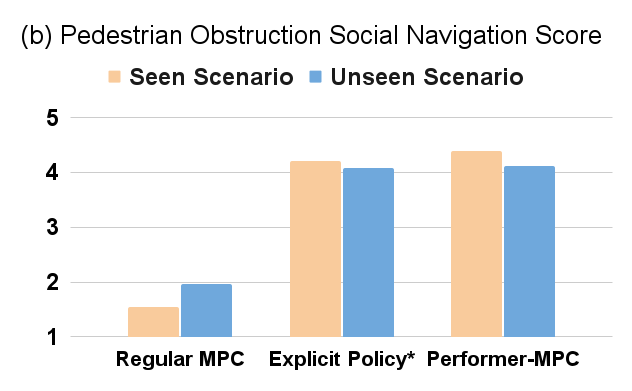}
\caption{Social Navigation Results. (a) In the \texttt{blind corner} scenario, \texttt{Performer-MPC} achieves a better social navigation score than \texttt{RMPC} and similar to \texttt{EP} in the \texttt{seen} environment. In the \texttt{unseen} environment \texttt{Performer-MPC} is better than both baselines. \texttt{EP} fails to complete \texttt{unseen} $20\%$ of the time (e.g., not reaching the goal or triggering a safety stop, denoted by *).  (b) In the \texttt{pedestrian obstruction} scenario, both \texttt{EP} and \texttt{Performer-MPC} show similar social scores and are superior to \texttt{RMPC}, but \texttt{EP} fails in \texttt{seen} or \texttt{unseen} conditions $5\%$ of the time (again denoted by *). \texttt{Performer-MPC} thus shows better generalizability in both social score and navigation success.}
\label{fig:results-social}
\vspace{-4mm}
\end{figure}

\vspace{-2mm}
\paragraph{Learning to Anticipate Pedestrians at Blind Corners: }
Going beyond static obstacles, we apply our method to social robot navigation \cite{mavrogiannis21-social-survey}, where robots must respect unwritten social norms for which cost functions are hard to design. One such scenario is \texttt{blind corner} (Fig.~\ref{fig::experiment}c), where robots should avoid the inner side of a hallway corner in case a human suddenly appears in this ``blind spot''. For \texttt{blind corner}, we collect 30 demonstrations with a human driving the robot from a randomly chosen location on one side of the corner to the other side in a socially compliant manner. After training, we evaluate each policy twenty times in the real world: ten times in the corner where the demonstrations were collected (\texttt{seen}), and ten times in a different corner (\texttt{unseen}). During each run, the robot and a \texttt{pedestrian} human subject will approach the corner from opposite sides, while a third-party \texttt{observer} monitors from a distance. We use a social navigation evaluation protocol \cite{pirk2022protocol} in which \texttt{pedestrian}s and \texttt{observer}s rate the performance of the policy using a standardized questionnaire  scored with Likert scales \cite{likert1932technique} which we combine into a joint social navigation score (see Appendix for further details). Social navigation scores for \texttt{blind corner} \texttt{seen} and \texttt{unseen} scenarios are shown in Fig. \ref{fig:results-social}a. \texttt{RMPC} has the least social compliance: its hand-crafted cost function efficiently cuts the corner, causing uncomfortable near-collisions (Fig.~\ref{fig:blind_corner}). \texttt{EP} performs slightly better than  \texttt{Performer-MPC} in the \texttt{seen} environments, but does not generalize well to \texttt{unseen} scenarios, with worse social scores and a $20\%$ failure rate (e.g., safety stops or not reaching the goal).
\vspace{-2mm}
\paragraph{Learning to Respect Comfort Distance When Obstructed by Pedestrians: }
Another common social navigation scenario is \texttt{pedestrian obstruction}, when a human unexpectedly impedes the prescribed path of a robot (Fig.~\ref{fig::experiment}d). While static obstacle avoidance is a largely solved problem, \texttt{pedestrian obstruction} is particularly challenging for MPC policies that are guided by waypoints that were valid before the human entered the environment. A hand-crafted cost function may guide the robot too near to the human, causing uncomfortably close interactions or the robot getting stuck right in front of the human. We evaluate policy performance for \texttt{pedestrian obstruction} using the social navigation evaluation protocol \cite{pirk2022protocol}.  Again, \texttt{Performer-MPC} is the most socially compliant (Fig.~\ref{fig:results-social}b), and in a few cases even shows emergent maneuvers unseen in the dataset (i.e., passing the human on the left if there is not enough space on the right, while the demonstrations only include right-side passing). In contrast, \texttt{RMPC} usually gets stuck in front of the human due to a local minimum close to the goal behind the human. While \texttt{EP} does stay away from the human subject in both \texttt{seen} and \texttt{unseen}, it struggles to reach the goal location and sometimes comes close to colliding with nearby walls, leading to a $5\%$ overall failure rate. Please refer to the Appendix for a detailed description of both social scenarios and statistical analysis of the results.

\subsection{Speed Studies Over Various Performer Architectures}
\label{sec:perablations}

Below we present speed ablation studies over various Performer architectures leading to Performer-ReLU as a default choice. More detailed studies are given in the Appendix.
Tests presented here are run on $100 \times 100$ images, with patch size $5 \times 5$ and two architecture sizes: (a) \textbf{medium} with $l=3$ layers, $h=1$ head, and $\mathrm{mlpdim}=64$ ($8.3\mathrm{M}$ parameters) and (b) \textbf{large} with  $l=6$ layers, $h=3$ heads, and $\mathrm{mlpdim}=1024$ ($24.6\mathrm{M}$ parameters).  We benchmark Performer-ReLU and Performer-Exp (for the latter one varying the number of random projections $\mathrm{rps}$ used and either redrawing them or not at each forward pass), measuring wall-clock time taken by the MPC. For the fastest Performer-ReLU variant, we run additional studies (this time measuring CPU-time to distill time taken solely by the MPC from I/O time, etc.) for varying patch sizes, showing that {\it we can reach Pixel-to-Pixel attention with near real-time speed} ($11.3\mathrm{ms}$ per MPC iteration). 
The results are presented in tables Tab. \ref{tab:main-abl-main} and \ref{tab:patch-abl-main}. 

\begin{table}[h]
\small
\centering
   \caption{Speed ablation tests for different variants of Performers with $100 \times 100$ input images. The architecture deployed on the real robot is denoted in \textbf{bold}.}
\label{tab:main-abl-main}
 \begin{tabular}{@{}lccccc@{}}
    \toprule
model size & Performer type & redraw & rps \# & wall-clock time per MPC-iteration [ms] \\
                   
\midrule
\textbf{medium} & \textbf{ReLU} & \textbf{N/A} & \textbf{N/A} & \textbf{8.3}\\
medium & Exp & False & 8 & 21.5\\
medium & Exp & False & 16 & 21.6\\
medium & Exp & False & 64 & 23.6\\

\midrule
large & ReLU & N/A & N/A & 13.8\\
large & Exp & False & 8 & 72.9\\
large & Exp & False & 16 & 66.9\\
large & Exp & False & 64 & 83.0\\
    \bottomrule
   \end{tabular}
    \vspace{-3mm}
\end{table}

\begin{table}[h]
\small
\centering
   \caption{Speed ablation over patch sizes for medium-sized Performer-ReLU with $100 \times 100$ inputs.}
\label{tab:patch-abl-main}
 \begin{tabular}{@{}lccccccc@{}}
    \toprule
patch sizes & 1 $\times \textrm{ }$ 1 & 2 $\times \textrm{ }$ 2 & 4 $\times \textrm{ }$ 4 & 5 $\times \textrm{ }$ 5 & 10 $\times \textrm{ } 10 $\\
                   
\midrule
number of params (M) & 15.7 & 9.93 & 8.50 & 8.33 & 8.16\\
CPU-time per MPC-iteration (ms) & 11.3 & 2.6 & 2.3 & 1.5 & 0.5\\
    \bottomrule
   \end{tabular}
  \vspace{-3mm}
\end{table}

%% file: conclusion.tex
\section{Limitations}
\label{sec:limitations}
Currently our Transformer-backend uses spatial attention, but in principle it can leverage the temporal axis. For example, in a face-to-face approach with a walking human in a hallway, motion history may shed light on how the human intends to move in the future, e.g., yielding left or right. A promising future research direction is to add history dependency so that the robot can sequentially reason about potential future interactions and conflicts. 
Furthermore, exploring richer modalities \cite{karnan2022socially} than the occupancy grid (e.g., RGB images, human traces, language contexts~\cite{ahn2022can,zeng2022socratic}), to enable robot-environment and robot-human interactions beyond simple geometry is another natural way to extend our approach. Another limitation is while the quadratic cost assures global convexity and training stability, it limits the expressiveness and complexity of the cost function. Furthermore, the cost function is learned individually for each navigation scenario, but it is unclear how one stand-alone learned cost function can handle multiple scenarios. Also, our user study pilot questionnaire could be refined, and our social evaluations could be expanded to a broader set of scenarios.

\vspace{-3.5mm}
\section{Conclusions}
\label{sec:conclusion}
\vspace{-2mm}
We present in this paper Performer-MPC, a learnable MPC system utilizing scalable Transformers to learn rich context representations parameterizing trainable cost function. 
We show that Performer-MPCs can be used as robotic controllers for navigation in challenging real-world environments where regular MPCs struggle, including learning to avoid local minima, to maneuver through highly constrained spaces, and to adhere to unwritten social norms, while maintaining real-time speed, even for nearly Pixel-to-Pixel attention.

%% file: supplementary.tex
\newpage
\appendix

\input{mpc_details}

\section{Speed studies over various Performers' architectures}
\label{sec:appendix_perablations}

We run speed studies over various Performer-MPC variants to choose the most suitable one for on-robot deployment characterized by strict latency constraints. The tests were run for $100 \times 100$ resolution images and two architecture sizes: large and medium (details below).

\paragraph{Large and medium architecture:} The large architecture consists of $l=6$ layers and $h=3$ heads with $\mathrm{mlpdim}=1024$. The medium architecture consists of $l=3$ layers and $h=1$ head with  $\mathrm{mlpdim}=64$. Both apply $\mathrm{GELU}$ nonlinearity in the MLP-layers. For $100 \times 100$ images, the large architecture has $24,581,732$ parameters, and the medium architecture has $8,334,884$ parameters.

\paragraph{Tested Performer variants:} We test $13$ different Performer versions: one Performer-ReLU and $12$ Performer-Exps. For Performer-Exps, we test both redrawing and no-redrawing, as well as a range of different random projections ($\mathrm{rps}$) including $8$, $16$, $32$, $64$, $128$, and $256$.

\paragraph{Results:} Our first set of results with patch size $5 \times 5$ is presented in Table \ref{tab:main-abl-app} (we report there wall-clock time). To obtain the desired speed $<0.1 \mathrm{sec}$ per MPC call with the average number of MPC iterations around $\mathrm{nb=10}$, we look for time per MPC-iteration $<10\mathrm{ms}$. Because this requirement is satisfied by the medium-size Performer-ReLU variant, we deploy it as our vision backend on-robot. We then study the inference speed of  this variant for different patch sizes and note that we can run it efficiently on robot with almost pixel-to-pixel ($1 \times 1$ patch size) attention resolution (for completeness, we include Table \ref{tab:patch-abl-main-app} here again, where we report CPU-times to accurately distill time taken by the MPC from other factors such as I/O time).

\begin{table}[h]
\small
\centering
   \caption{Full speed ablation tests for different variants of Performers with $100 \times 100$ input images. The architecture deployed on the real robot is denoted in \textbf{bold} (Extension of Tab. \ref{tab:main-abl-main}).}
\label{tab:main-abl-app}
 \begin{tabular}{@{}lccccc@{}}
    \toprule
model size & Performer type & redraw & nb rps & wall-clock time per MPC-iteration [ms] \\
                   
\midrule
\textbf{medium} & \textbf{ReLU} & \textbf{N/A} & \textbf{N/A} & \textbf{8.3}\\
medium & Exp & False & 8 & 21.5\\
medium & Exp & False & 16 & 21.6\\
medium & Exp & False & 32 & 22.2\\
medium & Exp & False & 64 & 23.6\\
medium & Exp & False & 128 & 24.7\\
medium & Exp & False & 256 & 26.8\\
medium & Exp & True & 8 & 73.6\\
medium & Exp & True & 16 & 67.6\\
medium & Exp & True & 32 & 74.0\\
medium & Exp & True & 64 & 75.0\\
medium & Exp & True & 128 & 79.0\\
medium & Exp & True & 256 & 81.6\\
\midrule
large & ReLU & N/A & N/A & 13.8\\
large & Exp & False & 8 & 72.9\\
large & Exp & False & 16 & 66.9\\
large & Exp & False & 32 & 77.4\\
large & Exp & False & 64 & 83.0\\
large & Exp & False & 128 & 83.0\\
large & Exp & False & 256 & 74.0\\
large & Exp & True & 8 & 104.1\\
large & Exp & True & 16 & 128.7\\
large & Exp & True & 32 & 116.2\\
large & Exp & True & 64 & 97.1\\
large & Exp & True & 128 & 143.5\\
large & Exp & True & 256 & 149.3\\
    \bottomrule
   \end{tabular}
\end{table}

\begin{table}[h]
\small
\centering
   \caption{Speed ablation over patch sizes for medium-sized Performer-ReLU with $100 \times 100$ inputs.}
\label{tab:patch-abl-main-app}
 \begin{tabular}{@{}lccccccc@{}}
    \toprule
patch sizes & 1 $\times \textrm{ }$ 1 & 2 $\times \textrm{ }$ 2 & 4 $\times \textrm{ }$ 4 & 5 $\times \textrm{ }$ 5 & 10 $\times \textrm{ } 10 $\\
                   
\midrule
number of params (M) & 15.7 & 9.93 & 8.50 & 8.33 & 8.16\\
CPU-time per MPC-iteration (ms) & 11.3 & 2.6 & 2.3 & 1.5 & 0.5\\
    \bottomrule
   \end{tabular}
\end{table}

\section{Experimental Setup}
We further provide details about our experimental setup. 

\subsection{Robot Setup}
Our robot can move at a linear speed between $[-0.8, 0.8]\textrm{m/s}$, and can rotate at an angular speed between $[-1.2, 1.2]\textrm{rad/s}$. The on-board software stack can perform SLAM (Simultaneous Localization And Mapping) and generate at each time step an occupancy grid with $0.05\times 0.05\textrm{m}$ cells (occupied or free) around the robot within $[-7.5, 7.5]\textrm{m}$ of range for both $x$ and $y$ axis. For most experiments such as in tight spaces and blind corners, we clip the occupancy grid to $[-2.5, 2.5]\textrm{m}$ ($100 \times 100$ cells) since there is enough local information to make navigation decisions. Only for the pedestrian obstruction scenario we reduce the occupancy grid to $[-4.5, 4.5]\textrm{m}$ ($180 \times 180$ cells) so the human can be detected early to leave enough decision making time. 

\subsection{Data Collection for Doorway Traversal}
\begin{figure}
\centering
\includegraphics[width=0.4\textwidth]{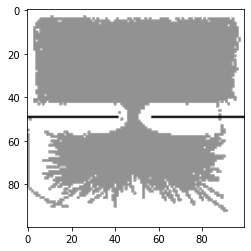}%
\caption{Expert Demonstrations Generated by an Off-Line MPC Planner Guided by a Dijkstra's Global Planner ($x$ and $y$ axis units are in occupancy grid cells). }
\label{fig::door_exprt_grey}
\end{figure}

To learn to avoid local minima in the doorway traversal scenario, we use an artificial expert to generate 2000 training episodes. Each episode is generated by (1) randomly sampling a start configuration on one side of the door and a goal position on the other side, (2) running a coarse Dijkstra's search to generate sparse global way points leading from the start through the doorway to the goal, and (3) feeding these way points on by one (i.e., using the solution of the last way point as the initial condition for the MPC solve for the next one) to an off-line MPC planner with 100 planning horizon and 200 maximum iterations (instead of 20 horizon and 20 iterations of the online MPC deployed on our robot). Such a heavy-duty planner requires too much computation to run onboard our robot. For the 2000 expert trajectories of 100 horizon, we apply a moving window of size 20 and extract 80 trajectories of 20 horizon, but keep the original goal of the 100-horizon trajectory on the other side of the wall, as expert demonstration data for the Performer-MPC to learn (the starting locations of these expert trajectories are shown in Fig. \ref{fig::door_exprt_grey}, with a few rare failure cases of the offline planner displayed by a few erratic points). 

\subsection{Data Collection for Highly Constrained Maneuvers}
To learn agile maneuvers in the highly constrained obstacle course, we collect data from a human expert via joystick teleoperation. The human expert teleoperates the robot to randomly navigate in a collision-free manner in the cluttered obstacle course (Fig. \ref{fig::experiment} b). The entire demonstration lasts around 30 minutes. For each data point, we set the goal as the 200-th future state on the demonstrated trajectory. At around $60\textrm{Hz}$ state rate and $0.5\textrm{m/s}$ driving speed, the goal is roughly $2\textrm{m}$ in front of the robot. In addition to this demonstration of desirable navigation behavior, we further collect about 10 minutes of demonstration starting from failure locations, e.g., where the robot is stuck and unable to recover from such situations, to address the distribution shift problem. 

\subsection{Data Collection for the Two Social Scenarios}
We defer the data collection details for the two social scenarios to Appendix~\ref{sec:social_study} after the social scenarios are formally defined to facilitate understanding. 

\subsection{Training Time}
Our bilevel optimization takes around 0.36 seconds per iteration, i.e., a forward inference pass and a backward training pass (Fig.~\ref{fig::overview}) to update the learnable parameters $\theta$ (Eqn.~\ref{eq:theta_star}) on four TPUs. As shown in Fig.~\ref{fig::test_data}, it takes around 15, 60, 1.5, and 10 hours for our Performer-MPC model to converge for the doorway traversal, highly constrained maneuvers, blind corner, and pedestrian obstruction scenarios, respectively.

\input{scenarios}

%% file: mpc_details.tex
\section*{Appendix}
\section{Differentiable MPC}
\label{sec:diff_mpc}

We provide some additional details regarding the learnable MPC, including the structure of the static cost and differentiation of the optimal solution w.r.t. the parameters of the cost neural networks.

\paragraph{Static Cost Function: }
Recall that the structure of our cost function inside the model predictive controller takes the form $\bar{c}(\x, \u, t) + c_{\mathrm{learn}}$ where $c_{\mathrm{learn}}(\x, \u; C_0, \theta)$ is the context-dependent residual cost. Let $\u := (\u_v, \u_{\omega})$ be the 2D control vector consisting of the body-aligned forward speed $\u_v$ and turn rate $\u_{\omega}$, and let $\x := (\mathbf{p}, \phi)$ be the state vector consisting of the 2D plane position $\mathbf{p}$ and orientation $\phi$. The static cost function takes the form:
\begin{subequations}
\begin{eqnarray}
    \bar{c}(\x, \u, t) &:= &\sum_{i=1}^{6} w_i \bar{c}_i(\x, \u, t) \\
    \bar{c}_1(\x, \u, t) &:= &\mathbb{I}_{\u_v \geq 0} |\u_v|^ 4, \quad \bar{c}_2(\x, \u, t) := \mathbb{I}_{\u_v < 0} |\u_v|^4, \quad \bar{c}_3(\x, \u, t) :=  |\u_{\omega}|^4\\
    \bar{c}_4(\x, \u, t) &:= & \sum_{j=1}^{n_c} \mathrm{ReLU}^2(\mathrm{margin} - d_j(\x))\\
    \bar{c}_5(\x, \u, t) &:= & \bar{w}_T(t) \|\mathbf{p} - \mathbf{g} \|^2, \qquad \bar{c}_6(\x, \u, t) := \bar{w}_T(t) (1 - \cos(\phi - g_{\phi}) ) \\
    \bar{w}_T(t) &:= &\begin{cases} 
        \dfrac{1}{T(1 + w_T)}  \quad &\text{if } t < T \\
        \dfrac{Tw_T + 1}{T(1 + w_T)} \quad &\text{if } t = T.
    \end{cases}
\end{eqnarray}
\end{subequations}
where $d_j(\x)$ is the signed distance from the $j^{th}$ collision-point on the robot body to the nearest obstacle, $\mathrm{margin}$ is a user-set margin threshold, $\mathbf{g} \in \mathbb{R}^2$ is a 2D goal position, $g_{\phi}$ is a goal orientation, and $\{w_i\}_{i=1}^{6}, w_T$ are a set of non-negative weights. Thus, the static/hand-engineered cost function consists of an asymmetric control penalty, margin-offset collision penalty, and a time-weighted goal-reaching penalty.

\paragraph{Tracking MPC: } The final policy action is determined as the solution to a secondary, non-learnable MPC problem (termed ``tracking MPC"), taking the form of~\eqref{eq:mpc} but with the cost comprising of only the static portion, i.e., $\bar{c}$. As defined above, this cost features a goal-reaching penalty. For the ``higher-level" MPC problem (i.e., the problem solved by either \texttt{Performer-MPC} or \texttt{RMPC}), this goal is determined from problem context, e.g., a global waypoint. For the ``tracking MPC" problem, this goal is set as an intermediate state extracted from the optimal state trajectory solution to the ``higher-level" MPC problem. In the \texttt{EP} case, this goal is directly output by the \texttt{EP} policy. Using such a dual/tracking-MPC structure allows a more fair comparison between the three policies since the lowest-level control action is output in an identical manner.

\paragraph{Jacobian of Optimal MPC solution: } The key tool for computing the desired Jacobian is the implicit function theorem, stated below (note: we suppress the dependence of the MPC cost function $J_c$ on the context $C_0$ for readability):

\begin{theorem}[Implicit Function Theorem] If in the neighborhood of $(\u^\star, \theta_k)$ where  $\nabla_{\u} J_c (\u^\star, \theta_k) = 0$, the Hessian $\nabla^2_{\u} J_c(\u^\star, \theta_k)$ is non-singular, then we have:
\begin{equation}
\partial_{\theta} \u^\star (\theta_k) = -\left[\nabla^2_{\u} J_c(\u^\star, \theta_k) \right]^{-1} \nabla^2_{\theta, \u} J_c(\u^\star, \theta_k) \label{eq:ift}
\end{equation}
\label{thm:ift}
\end{theorem}

The Hessian term above need not be explicitly materialized, since by chaining Eqns~\ref{eq:vjp1}~and~\ref{eq:ift}, the VJP can be efficiently calculated as,
\begin{equation}
 \nabla_{\theta} J_l(\mathbf{u}^{*}(\theta_k)) = -\left[\left[\nabla^2_{\u} J_c(\u^\star, \theta_k) \right]^{-1} \nabla_{\mathbf{u}} J_l(\mathbf{u}^{*}(\theta_k))\right]^T \nabla^2_{\theta, \u} J_c(\u^\star, \theta_k)  \label{eq:vjp2}
\end{equation}

The term inside the large square brackets may be computed as the solution to the following quadratic problem:
\[
    \argmin_{\delta \mathbf{v} := (\delta \mathbf{v}_0, \ldots, \delta \mathbf{v}_{T-1})} \quad \dfrac{1}{2} \delta \mathbf{v}^T \nabla^2_{\u} J_c(\u^\star, \theta_k) \delta \mathbf{v} + \delta \mathbf{v}^T \nabla_{\mathbf{u}} J_l(\mathbf{u}^{*}(\theta_k)),
\]
which in turn decomposes into a TV-LQR problem~\cite{dunn1989efficient} (Thm.~\ref{thm:ift}). Finally, the dot product with $\nabla^2_{\theta, \u} J_c(\u^\star, \theta_k)$ may be computed by differentiating the co-state equations associated with the gradient $\nabla_{\u} J_c(\u^\star, \theta_k)$.

%% file: scenarios.tex
\begin{figure}
\centering
  \centering
  \includegraphics[width=0.49\linewidth]{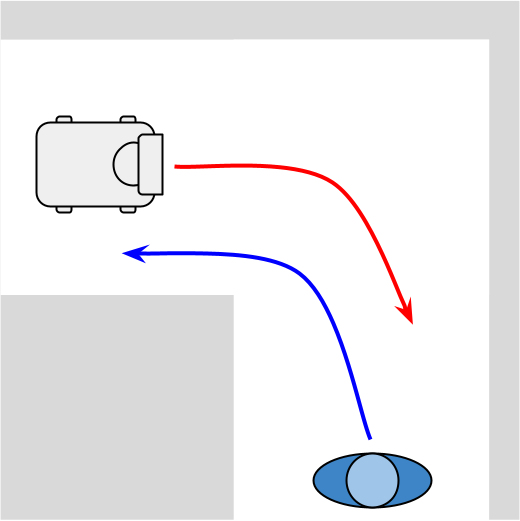}
  \centering
  \includegraphics[width=0.49\linewidth]{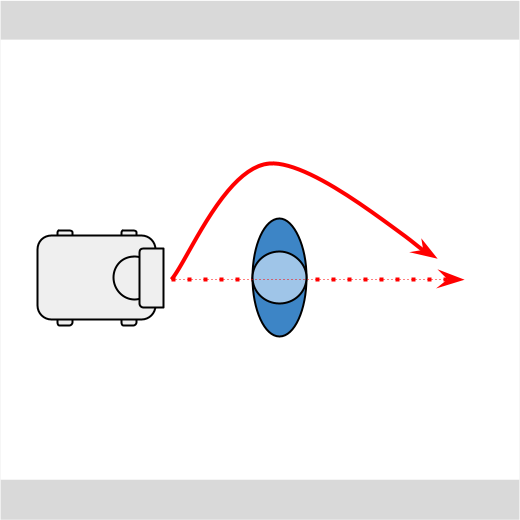}
\caption{Social Navigation Scenarios. (a) In the \texttt{blind corner} scenario, robots should swing wide or slow down to avoid possible collisions with approaching humans who are not visible. (b) In the \texttt{pedestrian obstruction} scenario, the robot should go around the visible obstructing human with a comfortable passing distance.}
\label{fig:social-scenarios}
\end{figure}

\begin{table} [t]
\small
\centering
\scalebox{0.95}{
  \begin{tabular}{l|l|l|l|l|l}
  \hline
  \multicolumn{2}{l|}{\textbf{Blind Corner}} & \multicolumn{2}{l|}{\textbf{Pedestrian Obstruction}} & \multicolumn{2}{l}{\textbf{General Questions}}  \\
  \hline
  BC1 & The robot moved & PO1 & The robot moved  & G1 & The robot adhered\\
  & to avoid me. & & to avoid me. &  & to social norms.\\
  BC2 & The robot stopped & PO2 & The robot maintained a safe & G2 & I would feel comfortable\\
    & to let me pass. & &  and comfortable distance. & &  around the robot in this\\
  BC3* & The robot nearly & PO3* & The robot nearly &  & encounter. \\
  & collided with me. &  & collided with me. & G3 & All things considered, I\\
  BC4* & I had to move & PO4 & It was clear what the &  & rate the robot motion as: \\ 
  & around the robot && robot wanted to do. &  & \texttt{Very Poor} / \texttt{Poor} / \texttt{Neutral} \\
  &  &  & & & / \texttt{Good} / \texttt{Very Good}. \\
  \hline   
  \end{tabular}
}
\vspace{2mm}
\caption{Questions for Each Social Scenario. A star indicates a question is negative, so that a `Strongly Disagree' result should be aligned with `Strongly Agree' for a positive question. 
}
\label{tab:questionnaire}
\end{table}

\section{Social Navigation Evaluation}
\label{sec:social_study}
To make evaluation of social navigation policies well-defined, realistic, scalable and repeatable, 
we use a social navigation benchmark based on a set of human-robot interaction scenarios to be evaluated using user surveys previously designed by \citet{pirk2022protocol}. This benchmark consists of scenarios with well-defined roles and expected behavior for both humans and robots, along with a series of questions for human raters with answers defined on a five-level Likert scale \cite{likert1932technique}. To provide a concise metric for comparing policies, we average the answers to these questions into a social navigation score reported in the main body of the paper (Fig.~\ref{fig:results-social}). This section justifies that choice by describing this social navigation protocol and the results of our experiments in more detail.

\subsection{Social Navigation Scenarios}

Crucially for our purposes, this protocol can be used to generate trajectories and evaluate whether they meet the scenario's social criteria, enabling us to create curated datasets of expert trajectories which score highly on our benchmark. For training Performer-MPCs, we chose two scenarios:

\begin{compactitem}
\item \texttt{blind corner} (Fig.~\ref{fig:social-scenarios}a), in which a robot is expected to apply some strategy (such as slowing down or swinging wide) to reduce the likelihood of collision with a possible unseen pedestrian coming around a corner.
\item \texttt{pedestrian obstruction} (Fig.~\ref{fig:social-scenarios}b), in which the robot is expected to drive around a human obstructing its path, while respecting the human's comfort distance. 
\end{compactitem}

Table \ref{tab:questionnaire} details the questions used to evaluate each scenario, along with a set of scenario-independent general questions. For most questions, the Likert scale is implemented as \texttt{Strongly Disagree, Disagree, Neutral, Agree, Strongly Agree}, except for question G3, which evaluates overall performance with \texttt{Very Poor, Poor, Neutral, Good, Very Good}.

\subsection{Collecting Expert Demonstrations}

We use the scenario definitions to collect a variety of expert demonstration datasets. For both scenarios, we collect around 30 episodes each scenario, which prove sufficient for training Performer-MPC. 
Note compare to taking visual RGB camera input which requires over 700 episodes each scenario~\cite{pirk2022protocol}, Performer-MPC takes occupancy grids as input and requires substantially smaller amount of data. 

For our social scenario data collection, human expert trajectories are collected by individuals trained as both participants and evaluators of the social scenarios, who attempt in their runs to guide the robot according to the social norms defined in the scenarios. In turn, our evaluation of Performer-MPC against baselines in these scenarios uses the same definitions and questionnaires which guide the trajectories; thus our experimental results gauge how well Performer-MPCs can successfully navigate with respect to social norms whose cost functions are difficult to explicitly design.

For \texttt{pedestrian obstruction}, we discover that both \texttt{Performer-MPC} and \texttt{EP} tend to memorize the building configurations of the training environment (e.g., walls, chairs, and tables) and sometimes do not respond to the human properly during deployment. Therefore, we augment the existing training data by randomly shuffling the background (i.e., randomly removing or adding obstacle pixels to the surrounding area, but keeping the space around the robot-human interaction point intact). We posit that such data augmentation may not be necessary when we scale up our data collection to different building configurations, and more importantly, adding extra information to distinguish humans from obstacles (e.g., human detection, tracking, and prediction). For the \texttt{blind corner} scenario, swinging wide to avoid the inner side of the corner is part of the desired learning process, therefore such augmentation is not necessary. 

We also randomly select goal locations between behind the human and the final robot position of each episode for the \texttt{pedestrian obstruction} scenario to improve the model's robustness against different goals. Again, for \texttt{blind corner}, we find such augmentation not necessary and simply select the 300-th future state on the demonstrated trajectory as goal for each data point. In this way, most data points have a goal behind the corner. We posit that these two data augmentation techniques contribute to the much longer training time for \texttt{pedestrian obstruction} than that for \texttt{blind corner} (Fig. \ref{fig::test_data}). 

\subsection{Human Evaluation Pilot Study}
To evaluate the performance of our social navigation policies, we conduct a pilot study, gathering both \texttt{participant} and \texttt{observer} perspectives using the scenario questionnaires. Due to covid-19 restrictions and limited availability of participants for in-person user study participation (N=9), we ran this pilot study with members of our research team. In this pilot study, we aim to address the research question: \textit{How well does our} \texttt{Performer-MPC} \textit{social navigation policy perform on our social navigation metrics when compared against} \texttt{RMPC} \textit{and} \texttt{EP}?

Beyond that research question, we also aim to explore several other variables. First, we want to see how these policies would perform when comparing previously seen environments (i.e., environments where the training set is collected) vs. unseen environments (i.e., novel environments not in the training set). Second, we want to examine how direct interactants (1st person) vs. bystanders (3rd person) would rate the robot’s social navigation performance because we hypothesize that 1st person responses might be stronger, especially when it comes to perceptions of comfort and safety. 
Third, we want to test the policies’ performances across at least two different social navigation scenarios, starting with \texttt{blind corner} and \texttt{pedestrian obstruction}. 

As this is a large set of variables (navigation policy, performing in seen vs. unseen environments, measuring from 1st vs. 3rd person perspectives, and navigating in two different social navigation scenarios) and we have a very limited set of research study participants, we opt to run this as an exploratory pilot study, not as a fully controlled, counterbalanced, human subjects experiment. Altogether we run 120 sessions, gathering 240 sets of questionnaire responses from 1st and 3rd person perspectives in the course of one day in June 2022 on our campus (N=9). To minimize possible bias, neither pedestrians nor observers are aware of what policy is being tested during each episode, and the ordering of policies is randomized.

\subsection{Evaluating Social Navigation Performance Factors}
To assess the relationship between people’s responses to our social navigation questionnaire items, we run principal component analyses (using varimax rotations) to see if the variables really do hang together. We also run reliability analyses to see if those items that load onto a single factor are indeed reliable measures of an underlying factor. 

\begin{compactitem}
\item \texttt{blind corner}: The three general questions in this set (Questions G1-G3) create a highly reliable factor, Cronbach’s alpha = .92 (N=120). When we run PCA on the questions that are specific to the \texttt{blind corner} scenario (Questions BC1-B4), one of the items does not correlate strongly with the other three (BC2: The robot stopped to let me pass).  
\item \texttt{pedestrian obstruction}: We find that the three general social navigation performance questions (Questions G1-G3) create a highly reliable factor, Cronbach’s alpha = .99 (N=240). For the four questions specific to the \texttt{pedestrian obstruction}, we find that all four questions (Questions PO1-PO4) also create a highly reliable factor, Cronbach’s alpha = .97 (N=120). 
\end{compactitem}

Because these results show most these variables are highly correlated, we average them into a ``social navigation score'' for presenting our results in the main body of the paper concisely. The following section presents the detailed results of the social questionnaire evaluation.

\begin{table}
\small
\centering
\scalebox{0.75}{
  \begin{tabular}{r|l||l|l|l|l||l|l|l||l|l|l}
  \hline
  \hline
   &  & BC1 & BC2 & BC3 & BC4 & G1 & G2 & G3 &  &  &  \\  Policy & Eval & Avoided & Stop for & No Fear & No & Social & User & Perceived & Social & Failed & Num \\
  Tested & Cond & Human & Human & Collision & Dodge & Norms & Comfort & Quality & Score & Episode & Samples \\
  \hline
  \hline
  Regular & Seen & 1.65 & 2.00 & 1.75 & 1.95 & 1.90 & 2.00 & 1.90 & 1.88 & 0\% & 60 \\
  MPC & \textit{Std. Err.} & \textit{0.08} & \textit{0.11} & \textit{0.14} & \textit{0.15} & \textit{0.12} & \textit{0.13} & \textit{0.12} & \textit{} & \textit{} & \textit{} \\
   &  &  &  &  &  &  &  &  &  &  &  \\
   & Unseen & 2.10 & 2.15 & 2.50 & 2.05 & 2.60 & 2.60 & 2.65 & 2.38 & 0\% & 60 \\
  \textit{} & \textit{Std. Err.} & \textit{0.09} & \textit{0.09} & \textit{0.11} & \textit{0.14} & \textit{0.11} & \textit{0.11} & \textit{0.10} & \textit{} & \textit{} & \textit{} \\
   &  &  &  &  &  &  &  &  &  &  &  \\
\hline
  Explicit & Seen & 3.90 & 1.90 & 4.65 & 4.70 & 4.65 & 4.70 & 4.60 & 4.16 & 0\% & 60 \\
  Policy & \textit{Std. Err.} & \textit{0.14} & \textit{0.14} & \textit{0.06} & \textit{0.06} & \textit{0.06} & \textit{0.06} & \textit{0.06} & \textit{} & \textit{} & \textit{} \\
   &  &  &  &  &  &  &  &  &  &  &  \\
   & Unseen & 3.40 & 1.80 & 4.00 & 3.70 & 2.95 & 3.25 & 3.10 & 3.17 & 20\% & 60 \\
  \textit{} & \textit{Std. Err.} & \textit{0.18} & \textit{0.14} & \textit{0.17} & \textit{0.20} & \textit{0.17} & \textit{0.18} & \textit{0.18} & \textit{} & \textit{} & \textit{} \\
   &  &  &  &  &  &  &  &  &  &  &  \\
\hline
  Performer- & Seen & 3.20 & 1.75 & 4.60 & 4.35 & 4.30 & 4.45 & 4.50 & 3.88 & 0\% & 60 \\
  MPC & \textit{Std. Err.} & \textit{0.17} & \textit{0.07} & \textit{0.10} & \textit{0.10} & \textit{0.08} & \textit{0.09} & \textit{0.09} & \textit{} & \textit{} & \textit{} \\
   &  &  &  &  &  &  &  &  &  &  &  \\
   & Unseen & 3.75 & 1.90 & 4.60 & 4.20 & 4.15 & 4.25 & 4.25 & 3.87 & 0\% & 60 \\
  \textit{} & \textit{Std. Err.} & \textit{0.13} & \textit{0.08} & \textit{0.06} & \textit{0.09} & \textit{0.10} & \textit{0.09} & \textit{0.08} & \textit{} & \textit{} & \textit{} \\
   &  &  &  &  &  &  &  &  &  &  &  \\
  \hline
  \hline
  \end{tabular}
}
\vspace{2mm}
\caption{Social Navigation Questionnaire Results for \texttt{blind corner}.}
\label{tab:social-results-blind-corner}
\end{table}

\begin{table}
\small
\centering
\scalebox{0.75}{
  \begin{tabular}{r|l||l|l|l|l||l|l|l||l|l|l}
  \hline
  \hline
   &  & PO1 & PO2 & PO3 & PO4 & G1 & G2 & G3 &  &  &  \\  
  Policy & Eval & Avoided & Comfort & No Fear & Motion & Social & User & Perceived & Social & Failed & Num \\
  Tested & Cond & Human & Distance & Collision & Legible & Norms & Comfort & Quality & Score & Episode & Samples \\
\hline
\hline
  Regular & Seen & 1.74 & 1.42 & 1.47 & 1.63 & 1.47 & 1.53 & 1.58 & 1.55 & 0\% & 60 \\
  MPC & \textit{Std. Err.} & \textit{0.13} & \textit{0.07} & \textit{0.08} & \textit{0.11} & \textit{0.07} & \textit{0.07} & \textit{0.07} & \textit{} & \textit{} & \textit{} \\
   &  &  &  &  &  &  &  &  &  &  &  \\
   & Unseen & 1.95 & 1.85 & 2.00 & 2.00 & 2.00 & 2.00 & 1.95 & 1.96 & 0\% & 60 \\
  \textit{} & \textit{Std. Err.} & \textit{0.21} & \textit{0.20} & \textit{0.19} & \textit{0.17} & \textit{0.19} & \textit{0.19} & \textit{0.19} & \textit{} & \textit{} & \textit{} \\
   &  &  &  &  &  &  &  &  &  &  &  \\
\hline
  Explicit & Seen & 4.45 & 4.15 & 4.40 & 3.80 & 4.15 & 4.30 & 4.15 & 4.20 & 5\% & 60 \\
  Policy & \textit{Std. Err.} & \textit{0.08} & \textit{0.12} & \textit{0.10} & \textit{0.12} & \textit{0.10} & \textit{0.10} & \textit{0.10} & \textit{} & \textit{} & \textit{} \\
   &  &  &  &  &  &  &  &  &  &  &  \\
   & Unseen & 4.30 & 4.15 & 4.05 & 3.80 & 4.10 & 4.10 & 4.05 & 4.08 & 5\% & 60 \\
  \textit{} & \textit{Std. Err.} & \textit{0.16} & \textit{0.15} & \textit{0.15} & \textit{0.17} & \textit{0.15} & \textit{0.15} & \textit{0.15} & \textit{} & \textit{} & \textit{} \\
   &  &  &  &  &  &  &  &  &  &  &  \\
\hline
  Performer- & Seen & 4.57 & 4.38 & 4.48 & 4.19 & 4.29 & 4.38 & 4.43 & 4.39 & 0\% & 60 \\
  MPC & \textit{Std. Err.} & \textit{0.07} & \textit{0.10} & \textit{0.10} & \textit{0.07} & \textit{0.12} & \textit{0.08} & \textit{0.08} & \textit{} & \textit{} & \textit{} \\
   &  &  &  &  &  &  &  &  &  &  &  \\
   & Unseen & 4.00 & 4.15 & 4.15 & 4.05 & 4.15 & 4.15 & 4.05 & 4.10 & 0\% & 60 \\
  \textit{} & \textit{Std. Err.} & \textit{0.19} & \textit{0.16} & \textit{0.16} & \textit{0.15} & \textit{0.16} & \textit{0.16} & \textit{0.15} & \textit{} & \textit{} & \textit{} \\
   &  &  &  &  &  &  &  &  &  &  &  \\
\hline
\hline
  \end{tabular}
}
\vspace{2mm}
\vspace{2mm}
\caption{Social Navigation Questionnaire Results for \texttt{pedestrian obstruction}.}
\label{tab:social-results-pedestrian-obstruction}
\end{table}

\subsection{Pilot Study Results}
As we are unable to fully balance the experiment design (e.g., getting each of our participants to try out each of the 24 experiment conditions once), we cannot satisfy the statistical analysis assumptions of repeated measures ANOVAs. As such we are reporting upon the descriptive statistics (means and standard errors) of our pilot study data, but we recommend interpreting these results as pilot study findings, not statistically significant findings that indicate causal relationships. 

\begin{compactitem}
\item \texttt{blind corner} (Tab.~\ref{tab:social-results-blind-corner}): The best performing policy on \texttt{blind corner} is the \texttt{EP} policy in the \texttt{seen} condition, with a combined score of $4.16$ and individual questionnaire scores equal to or higher than both other policies. However, \texttt{EP} does not generalize well to the \texttt{unseen} condition, suffering a $20\%$ failure rate and a drop in social score to $3.17$; differences between the performance of these policies on individual questions are generally greater than the standard errors of these policies, potentially indicating a real difference that could be teased out with a larger study. In contrast, \texttt{Performer-MPC} generalizes well, with scores on \texttt{seen} and \texttt{unseen} of $3.88$ and $3.87$ respectively, with individual questions generally showing greater differences. \texttt{RMPC} is the worst performing policy in the overall social score and on most individual questions, except BC2, ``The robot stopped to let me pass," on which it is slightly superior due to stopping for users. However, this illuminated issues  on question BC2, which we discuss further below. 
\item \texttt{pedestrian obstruction} (Tab.~\ref{tab:social-results-pedestrian-obstruction}): The best performing policy on \texttt{pedestrian obstruction} is \texttt{Performer-MPC} in the \texttt{seen} condition with overall social score of $4.39$, with a relatively small drop to $4.10$ in the \texttt{unseen} condition. \texttt{EP} also performs well with scores in \texttt{seen} and \texttt{unseen} of $4.20$ and $4.08$, respectively, though it fails to complete the task $5\%$ of the time in both conditions. While the difference in \texttt{seen} performance of these policies is typically greater than the standard error of their performance, potentially indicating a real difference which could be teased out with a larger study, this is not true in the \texttt{unseen} case. \texttt{RMPC} is the worst performing policy in both social score and individual questions. Results within questions, between questions and within policies under given conditions are generally more consistent for \texttt{pedestrian obstruction} than \texttt{blind corner}.
\end{compactitem}

Overall, we interpret these results to indicate that \texttt{Performer-MPC} has better generalization than  \texttt{EP} and is comparable in social navigation performance to \texttt{EP}, and that both of those policies are superior to \texttt{RMPC} at social navigation. The social navigation score results presented in the main body of the paper are consistent with this detailed analysis.

However, the outlier question BC2 is worth further discussion. All other questions in our survey hang together to create highly reliable factors, but BC2 does not, and show lower scores for both \texttt{Performer-MPC} and \texttt{EP}, which otherwise score highly on the social navigation questionnaire. Discussion with the participants and analysis of robot behaviors in the episodes reveal that this question inadvertently prescribes a solution: that the robot should stop at the blind corner. However, our expert training demonstrations incorporate a different solution: swinging wide at the corner to avoid collisions, which our human robot drivers determine is the preferred solution based on the navigation speed and stopping distance of the robot. In contrast, \texttt{RMPC}, while not scoring well on most social questions, nevertheless stop for the user, giving it an artificially high score on this question even though its behavior is not very social due to stopping very close to the user. While we report the results of this question for completeness, for future work we plan to craft questions which evaluate social navigation without prescribing a solution.

\subsection{Limitations and Future Work}

This paper focuses on whether Performer-MPC can successfully learn from expert demonstrations derived from real-world scenarios and then be successfully deployed in those scenarios on-robot. Therefore, we select a limited set of social scenarios which enables us to evaluate this research question. These scenarios are necessarily limited to those which can be detected from the occupancy grid, which preclude the use of the visual gesture-based scenarios proposed by \citet{pirk2022protocol}. Furthermore, data for these scenarios are collected by a limited number of human experts, and the policies for each scenario are trained separately. In future work, we plan to expand to a wider range of scenarios collected by a broader range of experts, and to train policies to solve sets of scenarios rather than a single scenario.

The user evaluation pilot study is a first step toward developing a more robust user study protocol for evaluating future versions of social navigation policies. Our pilot study is limited by its use of research team members as study participants; their perspectives on social navigation behavior are quite influenced by their experience with operating and running tests on these robots in their work. In the future, we will recruit user study participants from people, who are not part of our research team. Second, our pilot study is not properly balanced so we cannot run the usual statistical analyses necessary to evaluate the statistical significance of the effects we observed. Instead, we report upon the descriptive statistics for this paper and we will run a full user study as a next step in this research project. In future user studies, we will focus upon more targeted research questions so that the experiments are simpler to run, analyze, and interpret, and will ensure these questions focus on quality of social navigation without tying evaluation quality to mimicking a specific solution behavior.